\pdfoutput=1

\documentclass[11pt]{article}

\usepackage{emnlp2023}

\usepackage{times}
\usepackage{latexsym}
\usepackage{tabularx}
\usepackage{graphicx} 
\usepackage{multirow}
\usepackage{algorithm}
\usepackage{algpseudocode}
\usepackage{array}
\newcolumntype{C}{>{\centering\arraybackslash}X}
\usepackage{amsmath}
\usepackage{amssymb}
\usepackage{multirow}
\usepackage{array}
\usepackage{makecell}

\usepackage[T1]{fontenc}
\usepackage{ragged2e}
\usepackage{array}
\usepackage{adjustbox}
\usepackage{graphicx}
\usepackage{makecell}
\usepackage{xcolor}
\usepackage{booktabs}
\definecolor{darkgreen}{RGB}{0,100,0}
\usepackage{tikz}
\usetikzlibrary{positioning}
\usetikzlibrary{shapes.geometric}
\usetikzlibrary{arrows.meta}
\usepackage{pgfplots}
\pgfplotsset{compat=1.17}
\usepackage{graphicx}
\usepackage{caption}
\usepackage{multirow}
\usepackage[table]{xcolor}
\usepackage{array}
\usepackage{booktabs} 

\usepackage[utf8]{inputenc}

\usepackage{microtype}

\usepackage{inconsolata}

\title{DualFact+: A Multimodal Fact Verification Framework for Procedural Video Understanding}

\author{
Cennet Oguz, Yasser Hamidullah, Josef van Genabith \and Simon Ostermann \\
German Research Center for Artificial Intelligence (DFKI) \\
Saarland Informatics Campus \\
\texttt{\{cennet.oguz, yasser.hamidullah, josef.van\_genabith, simon.ostermann\}@dfki.de}
}

\begin{document}
\maketitle
\begin{abstract}
We introduce DualFact, a dual-layer, multimodal factuality evaluation framework for procedural video captioning. DualFact separates factual correctness into conceptual facts, capturing abstract semantic roles (e.g., Action, Ingredient, Tool, Location), and contextual facts, capturing their grounded predicate-argument realizations in video. To support complete and role-consistent evaluation, DualFact incorporates implicit argument augmentation (VIA) and contrastive fact sets. We instantiate DualFact in two modes: DualFact-T, which verifies facts against textual evidence, and DualFact-V, which verifies facts against video-grounded visual evidence. Experiments on YouCook3-Fact and CraftBench-Fact show that state-of-the-art multimodal language models produce fluent but often factually incomplete captions, with systematic omissions and role-level inconsistencies. DualFact correlates more strongly with human factuality judgments than standard metrics, particularly for contextual facts, and reveals that caption-only evaluation overestimates hallucinations compared to video-grounded verification. Overall, DualFact offers an interpretable and human-aligned evaluation protocol that highlights persistent challenges in multimodal factual grounding, extending beyond surface-level fluency.
\end{abstract}

\section{Introduction}

Procedural video captioning requires models to describe sequences of actions together with their associated arguments in a factual and visually grounded manner. While semantic role assignment is relevant across domains, it is particularly critical in procedural settings, where factual correctness depends not only on recognizing actions and objects, but also on assigning them to the correct roles (e.g., ingredient, tool, target). In instructional language, these roles are often implicit or underspecified (e.g., ``stir it''), creating ambiguity despite the referent being visually present in the video—a well-known challenge in procedural understanding~\cite{miech2019howto100m,zhou2018towards,kiddon2015mise,oguz2023find}.

Existing evaluation metrics fail to fully capture this challenge. Lexical metrics rely on surface overlap and cannot detect role or argument errors, while vision–language metrics measure global semantic similarity without structured reasoning over predicate–argument relations. Fact-based metrics verify extracted propositions but typically represent facts as flat, untyped statements, limiting their interpretability in procedural domains where correctness hinges on role assignment and step-level visual grounding~\cite{yuan2021video-fact,jiang2023hallucination,dziri2023faith}. To address these limitations, we introduce \textbf{DualFact}, a fact-based evaluation framework for procedural video captioning. DualFact represents each instructional step using a dual-layer fact structure that separates (1) contextual facts, capturing grounded predicate–argument relations visible in the video (e.g., \textit{cut(onion)}, \textit{stir(mixture, bowl)}), and (2) conceptual facts, capturing abstract role-level semantics such as \textsc{Action}, \textsc{Ingredient/Object}, \textsc{Tool}, and \textsc{Location}. This design enables fine-grained assessment of procedural factuality, distinguishing hallucinations, missing arguments, role inconsistencies, and visually grounded but task-irrelevant mentions in fluent yet incomplete captions.

Our contributions are threefold:  
(1) We introduce a dual-layer fact representation and a fact verification pipeline using textual and multimodal NLI models to assess factuality at both the grounded event and semantic role levels.  
(2) We construct two role-annotated factuality benchmarks\footnote{The code and data associated with this work are available at:
\url{https://github.com/OguzCennet/DualFact}.}, \textbf{YouCook3-Fact} and \textbf{CraftBench-Fact}, featuring clause-level segmentation, implicit argument augmentation, and structured fact annotations.  
(3) We show that DualFact reveals systematic omissions and role-level errors in state-of-the-art multimodal LLMs that are not captured by existing metrics.

\begin{table*}[ht]
\small
\renewcommand{\arraystretch}{1.25}
\resizebox{0.90\linewidth}{!}{
\begin{tabularx}{\textwidth}{p{0.1cm} p{1.5cm} p{7.0cm} p{5.0cm} p{2.7cm}}

& \textbf{Subtype} & \textbf{Error Definition} & \textbf{Reference Video} & \textbf{Caption} \\
\toprule

\multirow{3}{*}{\rotatebox[origin=c]{90}{\tiny {Hallucination}}} 
& Entity   
& Introduces an entity not appearing in the video. 
& Slicing \textcolor{darkgreen}{an onion} 
& slice \textcolor{red}{the garlic}. \\

& Predicate   
& Describes an absent action. 
& \textcolor{darkgreen}{Placing} bread on a plate 
& \textcolor{red}{toast} the bread. \\

& Tool/Location 
& Refers to an absent tool or location. 
& Mixing in \textcolor{darkgreen}{a pan} 
& mix in \textcolor{red}{a bowl}. \\
\midrule

\multirow{3}{*}{\rotatebox[origin=c]{90}
{\tiny{Salience}}} 
& Entity   
& Selects a visible but task-irrelevant entity. 
& Cutting \textcolor{darkgreen}{a cucumber} near a lemon 
& cut \textcolor{red}{the lemon}. \\

& Role 
& Swaps semantic roles among visible entities. 
& Adding \textcolor{darkgreen}{egg to milk} 
& add \textcolor{red}{milk to egg}.\\

& Tool/Location 
& Chooses a visible but irrelevant tool or location. 
& Stirring soup with \textcolor{darkgreen}{a spoon} in a pot 
& stir soup with \textcolor{red}{a ladle}. \\
\midrule

\multirow{2}{*}{\rotatebox[origin=c]{90}
{\tiny{Omission}}} 
& Entity     
& Omits a task-relevant entity. 
& Adding onion and \textcolor{darkgreen}{garlic} 
& add onion.\\

& Tool/Location 
& Omits a task-relevant tool or location. 
& Cutting a tomato with \textcolor{darkgreen}{a knife} on \textcolor{darkgreen}{a board} 
& cut tomato. \\
\bottomrule

\end{tabularx}
}
\caption{
Taxonomy of factual errors in video captioning. 
\textbf{Hallucination} introduces unsupported content, 
\textbf{Salience} misattributes visible but irrelevant elements, 
and \textbf{Omission} excludes task-relevant information.
}
\label{tab:error-taxonomy}
\end{table*}

\section{Related Work}

\subsection{Procedural Understanding}

Early video captioning models focused on generating fluent natural language descriptions from visual input~\cite{xu2016msr-vtt,zhou2018towards}. While large-scale pretrained multimodal models have improved visual grounding~\cite{chen2023internvl,miech2019howto100m}, they continue to struggle with procedural semantics, where captions must express explicit predicate--argument structure and resolve implicit arguments~\cite{kiddon2015mise,roth2015implicit}. Instructional datasets consistently reveal challenges such as missing arguments, role ambiguity, and multi-step dependencies, which limit the effectiveness of surface-level caption evaluation.

\subsection{Lexical and Vision--Language Metrics}

Lexical metrics such as BLEU~\cite{papineni2002bleu}, ROUGE~\cite{lin2004rouge}, and METEOR~\cite{denkowski-lavie-2014-meteor} quantify surface-level overlap but fail to capture role mismatches, argument swaps, or visually grounded omissions. Vision--language metrics, including CLIPScore~\cite{hessel2021clipscore}, EMScore~\cite{shi2022emscore}, PACScore~\cite{pacscore}, and UniEval~\cite{zhang2022unieval}, measure global semantic similarity in joint embedding spaces, but operate at a coarse granularity and lack structured representations of predicate--argument relations. As a result, these metrics are poorly suited to procedural settings where correctness depends on complete and role-consistent event structure.

\subsection{Multimodal Factuality and Grounding}

Recent work has introduced fact-based evaluation methods for image and video captioning~\cite{yuan2021video-fact,rohrbach2018object,jiang2023hallucination}. Approaches such as FaithScore~\cite{jing2023faithscore}, CapMAS~\cite{lee2024toward}, FactVC~\cite{liu2023models}, and FIFA~\cite{jing2025fifa} decompose captions into atomic statements and verify them using large vision--language models. While effective for identifying hallucinations and incorrect propositions, these methods typically represent facts as flat or weakly typed assertions, providing limited insight into role-specific errors, implicit argument recovery, or task-level completeness.

Our work builds on this line of research by adapting fact-level evaluation to the specific requirements of procedural video understanding. We introduce a dual-layer fact representation that separates grounded predicate--argument structure (\emph{contextual facts}) from abstract task-level semantics (\emph{conceptual facts}). This design enables structured reasoning about omissions, role inconsistencies, and visually grounded but task-irrelevant mentions, which are phenomena that prior factuality metrics do not explicitly model. In contrast to existing approaches, \textbf{DualFact} integrates role-aware semantics with multimodal grounding to support fine-grained factual assessment tailored to instructional videos.

\section{How-To Videos}

How-to videos provide dense multimodal supervision for learning real-world procedures across domains such as cooking, woodworking, and furniture assembly. They encode task knowledge through sequences of visual and verbal instructions, requiring captions that reflect both the physical execution of actions and the abstract semantic roles that structure each step.

\subsection{Contextual and Conceptual Facts}
\label{sec:dual_layer_facts}

Instructional steps express meaning at two complementary levels. We model this using conceptual and contextual facts.

Conceptual facts capture the abstract task semantics of a step, independent of its surface realization. They identify the intended action and its core participant roles (e.g., \textsc{ActionType}, \textsc{Ingredient/Object}, \textsc{Tool}, \textsc{Location}), normalizing across paraphrases such as “cut,” “slice,” or “prepare” the same ingredient. Contextual facts capture the visually grounded predicate--argument structure of the step as executed in the video, such as \textsc{cut}(\textit{tomato}, \textit{board}). They assess whether entities appear in the correct semantic roles during execution, distinguishing, for example, “pour water into flour” from “pour flour into water.” These layers frequently diverge: a caption may convey the correct conceptual step while misassigning arguments or roles at the contextual level. Explicitly separating abstract task meaning from grounded role realization is therefore essential for evaluating procedural captions, which exhibit high surface variability despite stable underlying structure.

\subsection{Error Taxonomy in Procedural Captions}
\label{sec:error_taxonomy}

Factual errors in procedural captions typically fall into three categories (Table~\ref{tab:error-taxonomy}). 
\emph{Hallucination} introduces content not supported by the video; 
\emph{Salience errors} incorrectly promote visible but task-irrelevant elements; and 
\emph{Omission} excludes task-critical information such as required entities, tools, or locations.  
These error types reflect failures at the conceptual layer, the contextual layer, or their interaction, motivating our dual-layer evaluation framework.

\begin{table}[t]
\centering
\small
\setlength{\tabcolsep}{4pt}
\begin{tabular}{llrrrrr}
\toprule
\textbf{Dataset} & \textbf{Split} & \textbf{Vid.} & \textbf{Clips} & \textbf{VIA} & \textbf{Con.} & \textbf{Ctx.} \\
\midrule
\multirow{2}{*}{YouCook3}
 & Train & 300 & 2{,}570 & 3{,}759 & 7{,}725 & 6{,}826 \\
 & Test  & 200 & 1{,}800 & 2{,}914 & 4{,}668 & 3{,}242 \\
\midrule
\multirow{2}{*}{CraftBench}
 & Train & 120 & 1{,}735 & 2{,}132 & 4{,}468 & 3{,}429 \\
 & Test  & 100 & 1{,}468 & 1{,}888 & 4{,}197 & 3{,}108 \\
\bottomrule
\end{tabular}
\caption{
Dataset statistics. VIA denotes annotated implicit arguments; Con. and Ctx. denote conceptual and contextual facts.
}
\label{tab:dataset-stats}
\end{table}

\section{Dataset and Annotation}

We construct two fact-level benchmarks for procedural video understanding: 
\textbf{YouCook3} for cooking videos and \textbf{CraftBench} for furniture crafting.
Both datasets provide clause-level captions, implicit argument augmentation, and structured factual annotations at \emph{contextual} and \emph{conceptual} levels.
Dataset statistics are shown in Table~\ref{tab:dataset-stats}, and full construction details are provided in Appendix~\ref{app:dataset}.

Existing instructional datasets such as HowTo100M~\cite{miech2019howto100m}, COIN~\cite{tang2019coin}, EPIC-KITCHENS~\cite{damen2018scaling}, and YouCook2~\cite{zhou2018towards} offer rich multimodal supervision but lack the structure required for factual evaluation, as captions often merge multiple actions, omit arguments, or underspecify semantic roles.
To address this, we standardize procedural captions through three steps:
(1) clause decomposition,
(2) implicit argument augmentation, and
(3) structured fact annotation.
Operational constraints and annotator-facing guidelines for each step are described in Appendix~\ref{app:annotator-instructions}.

\subsection{Caption Data and Clause Decomposition}

\paragraph{YouCook3.}
We extend YouCook2~\cite{zhou2018towards} by decomposing existing multi-action captions into atomic clauses, each aligned with a single action and a specific video segment.
For example, instructions such as “cut the tomatoes and potatoes and transfer to the pan” are split into separate verb-centered clauses with manually annotated temporal boundaries.
Details on clause atomicity, alignment criteria, and exclusion cases are provided in Appendix~\ref{app:clause}.

\paragraph{CraftBench.}
To evaluate cross-domain generality, we introduce \emph{CraftBench}, a new dataset of furniture and utility crafting videos involving tools and materials such as wood and metal.
For CraftBench, all instructional captions are annotated from scratch, rather than adapted from existing datasets.
Annotators produce concise, imperative-style captions guided by the narrator’s instructions and grounded in the visible execution, which are then temporally segmented and aligned with the corresponding action executions.
This yields fine-grained action–language correspondences in a mechanically oriented domain.
Dataset-specific annotation decisions and filtering criteria are described in Appendix~\ref{app:craftbench-annotation}.

Together, YouCook3 and CraftBench serve both as captioning benchmarks and as the basis for our factual annotation pipeline.

\subsection{Implicit Argument Augmentation}
\label{sec:implicit_annotation}

Procedural captions frequently omit arguments that are visually present or inferable from context~\cite{kiddon2015mise,roth2015implicit}.
We therefore annotate all missing but visually grounded arguments for \emph{patient}, \emph{tool}, and \emph{location} roles.
For example, “stir it” is augmented to “stir the soup with a spoon in the pot” when supported by the video.
This process yields the \textbf{YouCook3-VIA} and \textbf{CraftBench-VIA} variants, providing fully specified semantic representations of instructional steps.
The visual grounding criteria and role-coverage constraints used during VIA annotation are detailed in Appendix~\ref{app:via}.

\subsection{Atomic Fact Annotation}

From VIA-augmented captions, we derive two complementary fact layers per clause.
\emph{Contextual facts} (\(\mathcal{F}_r^{ctx}\)) encode grounded predicate–argument relations observable in the video (e.g., “add salt,” “add to the bowl”).
\emph{Conceptual facts} (\(\mathcal{F}_r^{con}\)) encode abstract role–value assignments such as \texttt{ACTION}, \texttt{INGREDIENT/OBJECT}, \texttt{TOOL}, and \texttt{LOCATION}.
Together, these layers support fine-grained factual verification across domains.
Annotation decisions, role definitions, and edge cases are described in Appendix~\ref{app:facts}.

\subsection{Negative Fact Construction}

To enable contrastive fact verification, each conceptual fact is paired with a semantically contradictory negative variant.
Negative facts are generated automatically by altering role values while preserving syntactic structure (e.g., replacing the correct tool or object with a plausible alternative).
These contrastive pairs supervise the distinction between supported and refuted facts in both textual and multimodal NLI-based verification.
Implementation details and the exact generation prompts are provided in Appendix~\ref{app:negfacts} and Appendix~\ref{app:negfact-prompts}.

\begin{figure*}[t]
    \centering
    \IfFileExists{fact_check_workflow.pdf}{%
    \includegraphics[width=0.99\textwidth]{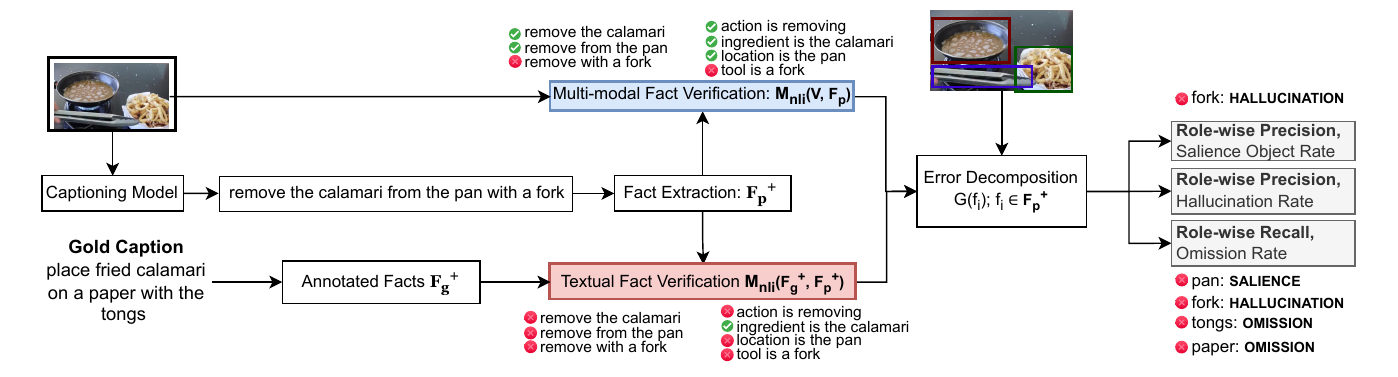}%
    }{%
    \fbox{\parbox{0.92\textwidth}{\centering
    Missing figure file: \texttt{emnlp2023-latex/pics/fact\_check\_workflow.pdf}.\\
    Upload the real workflow figure at this path before final submission.}}
    }
    \caption{Overview of the \textsc{MultiFactScore} pipeline. 
    The captioning model generates a description, from which positive facts are extracted and verified using multimodal and textual NLI models. 
    Error decomposition identifies hallucination, salience, and omission errors for fine-grained factuality analysis.}
    \label{fig:multifacts_pipeline}
\end{figure*}

\section{Method: MultiFactScore Framework}
\label{sec:method}

\textsc{MultiFactScore} evaluates factuality in procedural video captioning through two components:  
(1) extracting contextual and conceptual facts from both gold and generated captions, and  
(2) verifying these facts using multimodal or textual NLI models.  
This section describes fact construction, fact verification, and the final caption-level scoring procedure.

\subsection{Fact Generation}

For each instructional video segment, we derive three fact sets for both contextual
($\mathcal{F}^{ctx}$) and conceptual ($\mathcal{F}^{con}$) representations:
(i) \emph{positive gold facts} $\mathcal{F}_g^{+}$, manually annotated and grounded in the video;
(ii) \emph{negative gold facts} $\mathcal{F}_g^{-}$, automatically generated contradictions of $\mathcal{F}_g^{+}$;
and (iii) \emph{predicted facts} $\mathcal{F}_p$, extracted from the caption produced by a multimodal LLM.

\paragraph{Fact Extraction ($\mathcal{F}_p$).}  
Given a model-generated caption $\hat{C}$, we extract its atomic facts using \emph{Llama-3.3-70B-Instruct} through the Unsloth interface\footnote{\url{https://huggingface.co/unsloth/Llama-3.3-70B-Instruct}}:

\[
\mathcal{F}_p = \text{LLM}_{\text{extract}}(\hat{C}; \Phi)
= \{ f_1, \dots, f_n \}.
\]

Each fact corresponds to either a contextual predicate–argument structure (e.g., \textsc{add}(\textit{salt}, \textit{bowl})) or a conceptual role statement (e.g., \textsc{Location = bowl}). The fact extraction performance is evaluated in Appendix \ref{sec:eval_of_fact_extraction}.

\paragraph{Negative Fact Generation ($\mathcal{F}_g^{-}$).}  
To create contrastive supervision for NLI, each positive fact $f^{+} \in \mathcal{F}_g^{+}$ is paired with an incorrect but syntactically aligned variant: $\mathcal{F}_g^{-} = \text{LLM}_{\text{gen}}(\mathcal{F}_g^{+}; \Phi).$

Few-shot prompting encourages replacements of arguments (objects, tools, locations) while preserving the structure of the original fact.  
These negative facts are fluent and plausible, producing challenging refuted examples for verification.

\subsection{Fact Verification via NLI}

We use a natural language inference model $\mathcal{M}_{nli}$ to determine whether a fact $f_i$ is supported by the video $V$ (visual setting) or by the gold caption $C$ (textual setting).

\paragraph{Training (Multimodal Setting).}  
In multimodal verification, $\mathcal{M}_{nli}$ is trained on fact–video pairs:

\[
\hat{y}_i = \mathcal{M}_{nli}(V, f_i),
\qquad
\hat{y}_i \in \{\texttt{SUPPORTED}, \texttt{REFUTED}\}.
\]

\noindent Positive facts $\mathcal{F}_g^{+}$ are labeled \texttt{SUPPORTED} and negative facts $\mathcal{F}_g^{-}$ are labeled \texttt{REFUTED}.  
Textual NLI does not require training; we directly apply a pretrained LLM to judge textual entailment.

\paragraph{Evaluation.}  
At inference time, verification depends on the available evidence:

\[
\hat{y}_i =
\begin{cases}
\mathcal{M}_{nli}(V, f_i), & \text{visual verification}, \\[4pt]
\mathcal{M}_{nli}(C, f_i), & \text{textual verification}. \\
\end{cases}
\]

Ground truth labels follow the reference fact sets:

\[
y_i =
\begin{cases}
\texttt{SUPPORTED}, & f_i \in \mathcal{F}_g^{+}, \\
\texttt{REFUTED}, & f_i \in \mathcal{F}_g^{-}. \\
\end{cases}
\]

Classifier performance is measured using accuracy, precision, recall, and F1 over conceptual and contextual roles.

\paragraph{Per-Video Correctness.}  
For a video $v$, let $T(v)$ denote its set of fact types (e.g., \textsc{Action}, \textsc{Ingredient}, \textsc{Tool}).  
Accuracy is averaged across roles:

\[
\text{Acc}(v)
=
\frac{1}{|T(v)|}
\sum_{t\in T(v)}
\left(
\frac{1}{|t|}
\sum_{i\in t}
\mathbb{I}[\hat{y}_i = y_i]
\right).
\]

\subsection{Factuality of Generated Captions}

Given a video $V$, a captioning model produces $\hat{C} = \mathcal{M}_{\theta}(V)$.  
We extract its predicted facts $\mathcal{F}_p$ and evaluate them under two conditions:

\paragraph{Visual:} $\hat{y}_i = \mathcal{M}_{nli}(V, f_i),
\quad f_i \in \mathcal{F}_p.$

\paragraph{Textual:}  $\hat{y}_i = \mathcal{M}_{nli}(C, f_i),
\quad f_i \in \mathcal{F}_g^{+}.$

\paragraph{Caption-level MultiFactScore.}  
Let $F$ denote the relevant fact set ($\mathcal{F}_p$ for multimodal evaluation, $\mathcal{F}_g^{+}$ for textual).  
We compute:

\[
\text{MultiFactScore} =
\frac{
|\{ f_i \in F : \hat{y}_i = \texttt{SUPPORTED} \}|
}{
|F|
}.
\]

This score reflects the proportion of facts judged to be correct.

\subsection{Error Decomposition}

Beyond binary verification, we categorize factual inconsistencies into \emph{hallucinations}, \emph{salience errors}, and \emph{omissions}.  
For a predicted fact $f_i$, let $G(f_i)$ indicate whether the fact is visually grounded, computed using \emph{paligemma2-10b-pt-448}\footnote{\url{https://huggingface.co/google/paligemma2-10b-pt-448}}. The grounding performance is evaluated in Appendix \ref{sec:grounding_eval}.
\paragraph{Hallucination.}  
A fact is hallucinated if it is not visually grounded and the verifier labels it as \texttt{REFUTED}: $\text{Hallucination}(f_i)
=
\mathbf{1}[\neg G(f_i) \land f_i \in \mathcal{F}^{R}].$

\paragraph{Salience Error.}  
A fact is visually grounded but contradicted by the verifier: $\text{Salience}(f_i)
=
\mathbf{1}[G(f_i) \land f_i \in \mathcal{F}^{R}].$

\paragraph{Omission.}  
For any required entity $e_i$ in the gold fact set, omission occurs when the model fails to express it in any predicted fact: $\text{Omission}(e_i)
=
\mathbf{1}[\,e_i \in \mathcal{F}_g^{+} \land e_i \notin \mathcal{F}_p\,].$

This decomposition distinguishes unsupported predictions (hallucination), visually grounded but task-irrelevant mentions (salience), and missing task-critical facts (omission), providing stronger diagnostic insight into model behavior.

\begin{table*}[t]
\centering
\footnotesize
\setlength{\tabcolsep}{4pt}
\resizebox{0.75\linewidth}{!}{%
\begin{tabular}{lcccccccc}
\toprule
\textbf{} &
\textbf{VIA} &
\textbf{BLEU} &
\textbf{ROUGE} &
\textbf{SPICE} &
\textbf{BERTScore} &
$\texttt{\textbf{EMScore}}_{vc}$ &
$\texttt{\textbf{EMScore}}_{v}$ &
$\texttt{\textbf{EMScore}}_{c}$ \\
\midrule
\multirow{2}{*}{\rotatebox{90}{\textbf{CB}}}
& wo VIA & 1.17 & 18.00 & 7.47 & 86.93 & 51.12 & 27.92 & 74.33 \\
& w VIA  & 1.66 & 21.74 & 11.39 & 87.81 & 50.83 & 27.92 & 73.74 \\
\midrule
\multirow{2}{*}{\rotatebox{90}{\textbf{YC3}}}
& wo VIA & 5.87 & 24.16 & 21.26 & 87.14 & 52.84 & 28.01 & 77.67 \\
& w VIA  & 6.51 & 33.13 & 26.78 & 88.98 & 52.30 & 28.01 & 76.58 \\
\bottomrule
\end{tabular}
}
\caption{
NLG and EMScore results for CraftBench and YouCook3. “w VIA” denotes evaluation with implicit argument augmentation, and “wo VIA” denotes evaluation without it.
}

\label{tab:lexical_embedding_based_results_percent}
\end{table*}

\begin{table}[t]
\centering
\small
\begin{tabular}{lcccc}
\hline
\textbf{Dataset} & \textbf{Action} & \textbf{Ingredient} & \textbf{Tool} & \textbf{Location} \\
\hline
YouCook3    & 99.98 & 97.57 & 99.99 & 98.43 \\
CraftBench  & 98.65 & 96.48 & 99.52 & 96.74 \\
\hline
\end{tabular}
\caption{F1 scores for conceptual fact extraction on YouCook3-VIA and CraftBench-VIA.}
\label{tab:conceptual-fact-extraction}
\end{table}

\begin{table}[t]
\centering
\small
\begin{tabular}{lccccc}
\hline
\textbf{Dataset} & \textbf{act/obj} & \textbf{act/in} & \textbf{act/on} & \textbf{act/to} & \textbf{act/with} \\
\hline
YouCook3    & 98.35 & 99.35 & 99.21 & 99.46 & 100.0 \\
CraftBench  & 96.87 & 98.46 & 99.39 & 99.53 & 99.47 \\
\hline
\end{tabular}
\caption{F1 scores for contextual fact extraction on YouCook3-VIA and CraftBench-VIA.}
\label{tab:contextual-fact-extraction}
\end{table}

\begin{table*}[t]
\centering
\small
\setlength{\tabcolsep}{8pt}

\begin{tabular}{l l l c c c c c}
\toprule
\multirow{5}{*}{\rotatebox{90}{Conceptualized}}
& \textbf{Mode} & \textbf{Inputs} & \textbf{Action} & \textbf{Object}
& \textbf{Location} & \textbf{Tool} & \textbf{Avg.} \\
\cmidrule(l){2-8}
& Multimodal & $\mathcal{F}_g^{+}, \mathcal{F}_g^{-}, V$  & 92.50 & 81.53 & 90.50 & 86.30 & 88.07 \\
& Multimodal & $\mathcal{F}_p, V$    & 94.27 & 93.15 & 92.58 & 94.04 & 93.41 \\
\cmidrule(l){2-8}
& Textual    & $\mathcal{F}_g^{+}, \mathcal{F}_g^{-}, C$      & 98.81 & 99.06 & 99.02 & 98.77 & 98.92 \\
& Textual    & $\mathcal{F}_p,  C$     & 55.06 & 27.01 & 40.48 & 35.32 & 39.47 \\
\end{tabular}

\begin{tabular}{l l l c c c c c c}
\toprule
\multirow{5}{*}{\rotatebox{90}{Contextualized}}
& \textbf{Mode} & \textbf{Source} & \textbf{act/ing} & \textbf{act/in}
& \textbf{act/on} & \textbf{act/to} & \textbf{act/with} & \textbf{Avg.} \\
\cmidrule(l){2-9}
& Multimodal & $\mathcal{F}_g^{+}, \mathcal{F}_g^{-}, V$     & 78.68 & 83.43 & 80.35 & 82.67 & 77.80 & 79.89 \\
& Multimodal & $\mathcal{F}_p, V$    & 96.92 & 95.38 & 89.84 & 100.00 & 98.15 & 95.78 \\
\cmidrule(l){2-9}
& Textual    & $\mathcal{F}_g^{+}, \mathcal{F}_g^{-}, C$     & 95.05 & 93.66 & 92.63 & 94.36 & 93.45 & 93.83 \\
& Textual    & $\mathcal{F}_p, C$    & 16.72 & 20.52 & 19.76 & 29.21 & 21.92 & 21.23 \\
\bottomrule
\end{tabular}

\caption{
Unified fact-level evaluation on YouCook3-Fact. $\mathcal{F}_g^{+}$ denotes the annotated positive facts from the gold captions, while $\mathcal{F}_g^{-}$ refers to the generated negative facts constructed to contradict $\mathcal{F}_g^{+}$. $\mathcal{F}_p$ represents the facts extracted from the model-generated captions. $V$ and $C$ indicate the video clip and gold caption used as input for verification.
}

\label{tab:yc3_combined_all_facts}
\end{table*}

\begin{table}[h!]
\centering
\resizebox{0.90\linewidth}{!}{
\begin{tabular}{llccc}
\toprule
\small{\textbf{Fact Type}} & \small{\textbf{Eval Mode}} & \small{\textbf{Omission}} & \small{\textbf{Hallucination}} & \small{\textbf{Saliency}} \\
\midrule

\multirow{2}{*}{\small{\textbf{Ingredient}}} 
& cap-only     & 65.43 & 34.57 & -- \\
& cap-grounded   & 65.43 & 16.89 & 17.68 \\
& mm-grounded   & -- & 100.0 & 0.0 \\
\midrule

\multirow{2}{*}{\small{\textbf{Tool}}} 
& cap-only     & 49.80 & 50.20 & -- \\
& text-grounded   & 53.83 & 37.85 & 8.31 \\
& mm-grounded   & -- & 88.0 &  12.0\\
\midrule

\multirow{2}{*}{\small{\textbf{Location}}} 
& cap-only     & 40.03 & 59.97 & -- \\
& text-grounded   & 44.72 & 54.17 & 1.11 \\
& mm-grounded   & -- & 94.39 & 5.61 \\
\bottomrule
\end{tabular}
}
\caption{YouCook3-Fact error decomposition analysis. “Cap-only” uses caption-based verification only; “text-grounded” checks whether errors from textual verification are visually grounded; and “mm-grounded” checks whether errors from multimodal verification are visually grounded.}

\label{tab:yc3_error_decomp}
\end{table}

\section{Experiments}

We conduct three experiments on the YouCook3-Fact and CraftBench-Fact test splits to evaluate factuality under the \textsc{MultiFactScore} framework, using video clips, gold captions, and captions generated by the pretrained multimodal LLM \textbf{Qwen2.5-VL}.

\paragraph{Lexical and Vision--Language Metrics.}
We evaluate generated captions using standard lexical metrics (BLEU, ROUGE, SPICE, BERTScore) and EMScore, a vision--language metric based on joint embeddings.
We report three EMScore variants: $\textbf{EMScore}_{vc}$ (video + caption), $\textbf{EMScore}_{v}$ (video only), and $\textbf{EMScore}_{c}$ (caption only).

\paragraph{Verifier Performance.}
Before evaluating generated captions, we assess the multimodal and text-based NLI verifiers by pairing annotated positive facts $F_g^{+}$ and automatically generated negative facts $F_g^{-}$ with gold captions and video clips.
This experiment verifies the ability of the verifiers to distinguish supported from refuted facts in isolation.

\paragraph{Factual Evaluation of Generated Captions.}
We extract contextual ($\mathcal{F}_p^{ctx}$) and conceptual ($\mathcal{F}_p^{con}$) facts from model-generated captions and verify them against the corresponding video segments and gold annotations.
This evaluates end-to-end factual correctness and reveals omissions and role-level errors in generated captions.

\paragraph{Human Evaluation.}
We conduct a user study assessing factual correctness at the fact level.
Annotators judge whether each extracted fact is supported by (i) the gold caption (caption-based evaluation) and (ii) the video frames (video-based evaluation).
Facts are evaluated separately for conceptual roles (action, ingredient, tool, location) and contextual relations.
Annotators assign labels \emph{Correct}, \emph{Hallucination}, or (for video-based evaluation) \emph{Saliency}, capturing visually plausible but task-irrelevant content.

\section{Results and Discussion}

\begin{table*}[t]
\centering
\small
\setlength{\tabcolsep}{8pt}

\begin{tabular}{l l l c c c c c}
\toprule
\multirow{5}{*}{\rotatebox{90}{Conceptualized}}
& \textbf{Mode} & \textbf{Inputs} & \textbf{Action} & \textbf{Object}
& \textbf{Location} & \textbf{Tool} & \textbf{Avg.} \\
\cmidrule(l){2-8}
& Multimodal & $\mathcal{F}_g^{+}, \mathcal{F}_g^{-}, V$     & 81.14 & 80.22 & 78.70 & 77.66 & 79.93 \\
& Multimodal & $\mathcal{F}_p,V$      & 85.22 & 87.97 & 91.35 & 68.80 & 83.33 \\
\cmidrule(l){2-8}
& Textual    & $\mathcal{F}_g^{+}, \mathcal{F}_g^{-}, C$     & 96.65 & 90.78 & 98.77 & 95.08 & 94.82 \\
& Textual    & $\mathcal{F}_p,C$     & 20.24 & 13.57 & 13.76 & 28.11 & 19.92 \\
\end{tabular}

\begin{tabular}{l l l c c c c c c}
\toprule
\multirow{5}{*}{\rotatebox{90}{Contextualized}}
& \textbf{Mode} & \textbf{Inputs} & \textbf{act/obj} & \textbf{act/in}
& \textbf{act/on} & \textbf{act/to} & \textbf{act/with} & \textbf{Avg.} \\
\cmidrule(l){2-9}
& Multimodal & $\mathcal{F}_g^{+}, \mathcal{F}_g^{-}, V$     & 76.87 & 78.14 & 70.67 & 74.29 & 81.61 & 78.29 \\
& Multimodal & $\mathcal{F}_p,V$     & 92.13 & 97.06 & 91.56 & 85.71 & 89.12 & 91.12 \\
\cmidrule(l){2-9}
& Textual    & $\mathcal{F}_g^{+}, \mathcal{F}_g^{-}, C$     & 98.40 & 96.96 & 96.46 & 98.11 & 97.76 & 97.95 \\
& Textual    & $\mathcal{F}_p,C$    & 17.69 & 13.89 & 23.50 & 15.38 & 25.80 & 20.76 \\
\bottomrule
\end{tabular}

\caption{
Unified fact-level evaluation on CraftBench-Fact. $\mathcal{F}_g^{+}$ denotes the annotated positive facts from the gold captions, while $\mathcal{F}_g^{-}$ refers to the generated negative facts constructed to contradict $\mathcal{F}_g^{+}$. $\mathcal{F}_p$ represents the facts extracted from the model-generated captions. $V$ and $C$ indicate the video clip and gold caption used as input for verification.
}
\label{tab:ww_combined_all_facts}
\end{table*}

\begin{table}[h!]
\centering

\resizebox{0.90\linewidth}{!}{
\begin{tabular}{llccc}
\toprule
\small{\textbf{Fact Type}} & \small{\textbf{Eval Mode}} &
\small{\textbf{Omission}} & \small{\textbf{Hallucination}} & \small{\textbf{Saliency}} \\
\midrule

\multirow{2}{*}{\small{\textbf{Object}}} 
& cap-only     & 51.06 & 48.94 & -- \\
& cap-grounded   & 51.06 & 12.77 & 36.17 \\
& mm-grounded   & -- & 82.68 &  17.32\\
\midrule

\multirow{2}{*}{\small{\textbf{Tool}}} 
& cap-only     & 60.02 & 39.98 & -- \\
& cap-grounded   & 60.02 & 37.04 & 2.94 \\
& mm-grounded   & -- &  97.08 &  2.92 \\
\midrule

\multirow{2}{*}{\small{\textbf{Location}}} 
& cap-only     & 16.79 & 83.21 & -- \\
& cap-grounded   & 16.92 & 58.02 & 25.06 \\
& mm-grounded   & -- & 100.0 & 0.0 \\
\bottomrule
\end{tabular}
}
\caption{CraftBench-Fact error decomposition analysis. “Cap-only” uses caption-based verification only; “text-grounded” checks whether errors from textual verification are visually grounded; and “mm-grounded” checks whether errors from multimodal verification are visually grounded.}
\label{tab:craftbench_error_decomp}
\end{table}

\subsection{Lexical and Embedding-Based Evaluation}
Table~\ref{tab:lexical_embedding_based_results_percent} shows that implicit argument augmentation (VIA) consistently improves lexical caption quality for both CraftBench and YouCook3. VIA yields higher BLEU, ROUGE, SPICE, and BERTScore, indicating better alignment with the gold captions. In contrast, EMScore remains stable or decreases slightly, suggesting that the more explicit captions produced with VIA introduce structural changes that lexical metrics reward but embedding-based metrics treat as reduced semantic similarity. Gains are larger on YouCook3, reflecting the greater benefit of argument completion in more complex procedural descriptions.

\subsection{Fact Extraction Evaluation}
\label{sec:eval_of_fact_extraction}

We evaluate fact extraction by comparing model-generated structured facts with gold tuples from the test set.  
Each predicted slot value is treated as a binary correctness decision relative to its gold annotation.  
We compute precision, recall, and F1 at the slot level for both conceptual roles (e.g., \texttt{action}, \texttt{ingredient}) and contextual relations (e.g., \texttt{act/in}, \texttt{act/on}, \texttt{act/with}).  
F1 is reported as the primary performance metric.

The resulting F1 scores, shown in Tables~\ref{tab:conceptual-fact-extraction} and \ref{tab:contextual-fact-extraction}, indicate consistently high extraction accuracy across both YouCook3-Fact and CraftBench-Fact test sets.  
F1 values typically exceed 96--99\%, demonstrating that the LLM-based extractor reliably captures action semantics, entities, and role assignments from captions.  
This strong extraction performance suggests that downstream factuality evaluation is not dominated by extraction noise.

To quantify the impact of extraction errors on the final factuality score, we performed a sensitivity analysis on a YouCook3-Fact and CraftBench-Fact test sets. Replacing extracted facts with gold facts changed the final MultiFactScore by less than 0.8 points on average for conceptual roles and 1.2 points for contextual relations. This indicates that extraction noise does not materially affect downstream factuality evaluation.

\subsection{YouCook3-Fact Results}
Two main observations emerge from YouCook3-Fact:  
(1) The verifier performs well on gold facts, but predicted captions often omit or distort task-relevant information;  
(2) caption-only evaluation inflates hallucination rates, while visual grounding exposes saliency errors and corrects misclassifications.  
These findings highlight the necessity of video-grounded, fact-level evaluation.

\paragraph{Multimodal and Textual Fact Verification.}
Table~\ref{tab:yc3_combined_all_facts} reports conceptual and contextual fact verification results. Using gold conceptual facts, the multimodal verifier achieves strong accuracy (88.07\%), while contextual relations are more challenging (79.89\%), especially those involving fine-grained prepositions (\textit{act/on}, \textit{act/with}). Accuracy rises substantially when verifying facts extracted from predicted captions: 93.41\% for conceptual roles and 95.78\% for contextual relations, with some categories (e.g., \textit{act/to}) reaching 100\%. This pattern indicates a model–model consistency bias: the verifier aligns more easily with the captioning model’s own factual outputs than with the gold annotations.

\paragraph{Factual Correctness of Predicted Captions.}
When classifying gold facts relative to gold captions, the verifier achieves near-ceiling performance (98.92\% conceptual; 93.83\% contextual). However, when verifying predicted captions against the gold fact sets, accuracy drops sharply to 39.47\% for conceptual roles and 21.23\% for contextual relations. These results show that the captioning model generates fluent descriptions but fails to encode many of the task-relevant entities, relations, and role structures present in the instructional ground truth.

\paragraph{Error Decomposition.}
Table~\ref{tab:yc3_error_decomp} decomposes errors into omission, hallucination, and saliency. Caption-only evaluation consistently overestimates hallucination, since any mismatch with the gold facts is treated as unsupported. For instance, tools exceed 50\% hallucination and locations approach 60\%. Visual grounding reveals that many of these cases are saliency errors. For ingredients, hallucination drops from 34.57\% to 16.89\%, with 17.68\% reclassified as saliency; tools and locations follow the same trend. A persistent failure mode remains for actions: under multimodal grounding, 100\% of action-related errors remain hallucinations, indicating deeper semantic failures not recoverable from visual cues.

\subsection{CraftBench-Fact Results}

\paragraph{Multimodal and Textual Fact Verification.}
Table~\ref{tab:ww_combined_all_facts} shows strong verifier performance on CraftBench gold facts. Using gold conceptual facts, multimodal accuracy reaches 79.93\%, and contextual relations reach 78.29\%. Textual verification is near perfect (94.82\% conceptual; 97.95\% contextual). As in YouCook3, verification becomes easier when using predicted caption facts: 83.33\% accuracy for conceptual roles and 91.12\% for contextual relations, again reflecting a model–model consistency bias.

\paragraph{Factual Correctness of Predicted Captions.}
Despite strong performance on gold facts, factual correctness drops substantially when verifying predicted captions against gold annotations. Textual verification is particularly low, at 19.92\% accuracy for conceptual roles and 20.76\% for contextual relations. This indicates that the generated captions omit many task-relevant entities and relations found in the gold annotations, even though the verifier itself is strong.

\paragraph{Error Decomposition.}
Table~\ref{tab:craftbench_error_decomp} shows the same trends as YouCook3: caption-only evaluation greatly overestimates hallucination, e.g, 48.94\% objects; 39.98\% tools; 83.21\% locations. Visual grounding reclassifies many of these errors as saliency, entities visible in the scene but not part of the gold semantics, such as 36.17\% for objects and 25.06\% for locations. Under multimodal grounding, the remaining errors are genuine hallucinations, e.g., 82.68\% for objects; 97.08\% for tools; 100\% for location, revealing persistent challenges in grounding spatial and tool-related predictions even with video evidence.
\subsection{User Study Results}

Table~\ref{tab:human-eval-conctx} reports the distribution of human factual correctness judgments across evaluation settings, decomposed by modality (caption vs.\ video) and fact abstraction level (conceptual vs.\ contextual).
Caption-based evaluation is substantially stricter than video-based evaluation, with less than half of conceptual facts and only 21.55\% of contextual facts judged as correct.
In contrast, video-based evaluation yields markedly higher correctness rates for both conceptual (84.30\%) and contextual (60.34\%) facts, while also introducing saliency-based judgments that reflect visually plausible but non-essential information.

\begin{table}[t]
\centering
\small
\begin{tabular}{lccc}
\toprule
\textbf{Setting} &
\textbf{Correct} &
\textbf{Hallucination} &
\textbf{Saliency } \\
\midrule
Caption--Con & 45.66 & 54.34 & -- \\
Caption--Ctx & 21.55 & 78.45 & -- \\
\midrule
Video--Con & 84.30 & 8.14 & 7.56 \\
Video--Ctx & 60.34 & 21.55 & 18.10 \\
\bottomrule
\end{tabular}
\caption{
Distribution of human factual correctness judgments across evaluation settings.
Caption-based evaluation is substantially stricter, particularly for contextual facts, whereas video-based evaluation yields higher correctness rates and introduces saliency-based judgments.
}
\label{tab:human-eval-conctx}
\end{table}

We next examine how these modality- and abstraction-dependent human judgments align with automatic evaluation metrics.
Table~\ref{tab:userstudy-correlation} reports correlation results between human judgments and a range of standard NLG metrics, EMScore variants, and our fact-based evaluation scores.
Among all methods, caption-based conceptual fact scores achieve the strongest agreement with human judgments across all three correlation measures, with Spearman’s $\rho$ reaching 0.429.
Contextual fact scores yield consistently lower correlation, while video-based fact scores exhibit weaker rank-based agreement overall.

\begin{table}[t]
\centering
\small
\begin{tabular}{lccc}
\toprule
\textbf{Method} & \textbf{Pearson $r$} & \textbf{Spearman $\rho$} & \textbf{Kendall $\tau$} \\
\midrule
CIDEr & 0.138 & 0.140 & 0.093 \\
BERTScore & 0.047 & -0.050 & -0.031 \\
\midrule
EMScore (Text) & 0.327 & 0.280 & 0.207 \\
EMScore (Video) & 0.312 & 0.019 & 0.006 \\
\midrule

Caption (Con) & \textbf{0.534} & \textbf{0.429} & \textbf{0.337} \\
Caption (Ctx) & 0.460 & 0.368 & 0.297 \\
Video (Con) & 0.423 & 0.279 & 0.232 \\
Video (Ctx) & 0.238 & 0.222 & 0.186 \\
\bottomrule
\end{tabular}
\caption{
Correlation between automatic evaluation metrics and human factual correctness judgments from the user study.
Spearman’s $\rho$ is reported as the primary measure of agreement.
Best results are shown in bold.
}
\label{tab:userstudy-correlation}
\end{table}

Although video-based human judgments exhibit higher absolute correctness (Table~\ref{tab:human-eval-conctx}), automatic video-based fact scores are likewise saturated at high values.
This leads to ceiling effects and reduced score variance, which in turn weakens rank-based correlation with human judgments, despite stronger absolute agreement.
In contrast, caption-based conceptual judgments produce a more discriminative range of scores, resulting in stronger alignment with automatic fact-based evaluation.

\section{Conclusion}

We introduced MultiFactScore, a role-aware factuality framework tailored to procedural video captioning. Unlike prior fact-based metrics, MultiFactScore separates conceptual role semantics from contextual predicate–argument grounding, enabling fine-grained assessment of omissions, role inconsistencies, and visually grounded but task-irrelevant mentions. Experiments on YouCook3-Fact and CraftBench-Fact highlight persistent factuality challenges in current multimodal captioning models and demonstrate the value of separating conceptual meaning from visual realization. The framework establishes a foundation for future systems that aim to generate more robust, role-consistent, and visually aligned procedural descriptions.

Overall, \textsc{MultiFactScore} provides an interpretable and empirically grounded approach to factuality assessment in instructional video captioning. The dual-layer fact representation and fine-grained error decomposition offer a foundation for future work on generating factually grounded, role-consistent, and visually aligned procedural descriptions.

\section{Limitations}

Our work has several limitations. First, the benchmarks are restricted to cooking and furniture crafting videos, which may limit the generalizability of our methods to other domains without further adaptation. Second, our fact-based evaluation relies on the accuracy of the underlying fact extraction pipeline; any errors in extraction propagate to the final evaluation. Third, we focus exclusively on action- and object-related facts, and do not model attribute-oriented facts, such as size or spatial properties. Fourth, video-based fact checking remains challenging, particularly in visually complex scenes involving occlusions or fine-grained spatial relationships, which can reduce accuracy. Finally, although our approach identifies hallucinated facts, it does not yet distinguish between different types or severities of hallucination.

\section*{Acknowledgments}
This work was supported by the European Union through the PERKS project, 
``Eliciting and Exploiting Procedural Knowledge in Industry 5.0'' 
(Grant Agreement No. 101120323). We also thank Yana Veitsman, 
Nellia Dzhubaeva, and Bangyao Tang for their help with the annotation.


\bibliography{anthology,custom}

@inproceedings{zhou2018towards,
  title={Towards automatic learning of procedures from web instructional videos},
  author={Zhou, Luowei and Xu, Chenliang and Corso, Jason},
  booktitle={Proceedings of the AAAI conference on artificial intelligence},
  volume={32},
  number={1},
  year={2018}
}

@inproceedings{papineni2002bleu,
  title={Bleu: a method for automatic evaluation of machine translation},
  author={Papineni, Kishore and Roukos, Salim and Ward, Todd and Zhu, Wei-Jing},
  booktitle={Proceedings of the 40th annual meeting of the Association for Computational Linguistics},
  pages={311--318},
  year={2002}
}

@inproceedings{lin2004rouge,
  title={Rouge: A package for automatic evaluation of summaries},
  author={Lin, Chin-Yew},
  booktitle={Text summarization branches out},
  pages={74--81},
  year={2004}
}

@article{hessel2021clipscore,
  title={Clipscore: A reference-free evaluation metric for image captioning},
  author={Hessel, Jack and Holtzman, Ari and Forbes, Maxwell and Bras, Ronan Le and Choi, Yejin},
  journal={arXiv preprint arXiv:2104.08718},
  year={2021}
}

@inproceedings{shi2022emscore,
  title={Emscore: Evaluating video captioning via coarse-grained and fine-grained embedding matching},
  author={Shi, Yaya and Yang, Xu and Xu, Haiyang and Yuan, Chunfeng and Li, Bing and Hu, Weiming and Zha, Zheng-Jun},
  booktitle={Proceedings of the IEEE/CVF conference on computer vision and pattern recognition},
  pages={17929--17938},
  year={2022}
}

@article{liu2023models,
  title={Models see hallucinations: Evaluating the factuality in video captioning},
  author={Liu, Hui and Wan, Xiaojun},
  journal={arXiv preprint arXiv:2303.02961},
  year={2023}
}

@inproceedings{miech2019howto100m,
  title={Howto100m: Learning a text-video embedding by watching hundred million narrated video clips},
  author={Miech, Antoine and Zhukov, Dimitri and Alayrac, Jean-Baptiste and Tapaswi, Makarand and Laptev, Ivan and Sivic, Josef},
  booktitle={Proceedings of the IEEE/CVF international conference on computer vision},
  pages={2630--2640},
  year={2019}
}

@inproceedings{tang2019coin,
  title={Coin: A large-scale dataset for comprehensive instructional video analysis},
  author={Tang, Yansong and Ding, Dajun and Rao, Yongming and Zheng, Yu and Zhang, Danyang and Zhao, Lili and Lu, Jiwen and Zhou, Jie},
  booktitle={Proceedings of the IEEE/CVF Conference on Computer Vision and Pattern Recognition},
  pages={1207--1216},
  year={2019}
}

@inproceedings{kiddon2015mise,
  title={Mise en place: Unsupervised interpretation of instructional recipes},
  author={Kiddon, Chlo{\'e} and Ponnuraj, Ganesa Thandavam and Zettlemoyer, Luke and Choi, Yejin},
  booktitle={Proceedings of the 2015 Conference on Empirical Methods in Natural Language Processing},
  pages={982--992},
  year={2015}
}

@article{chen2023internvl,
  title={InternVL: Scaling up Vision Foundation Models and Aligning for Generic Visual-Linguistic Tasks},
  author={Chen, Zhe and Wu, Jiannan and Wang, Wenhai and Su, Weijie and Chen, Guo and Xing, Sen and Zhong, Muyan and Zhang, Qinglong and Zhu, Xizhou and Lu, Lewei and Li, Bin and Luo, Ping and Lu, Tong and Qiao, Yu and Dai, Jifeng},
  journal={arXiv preprint arXiv:2312.14238},
  year={2023}
}

@inproceedings{denkowski-lavie-2014-meteor,
    title = "Meteor Universal: Language Specific Translation Evaluation for Any Target Language",
    author = "Denkowski, Michael  and
      Lavie, Alon",
    editor = "Bojar, Ond{\v{r}}ej  and
      Buck, Christian  and
      Federmann, Christian  and
      Haddow, Barry  and
      Koehn, Philipp  and
      Monz, Christof  and
      Post, Matt  and
      Specia, Lucia",
    booktitle = "Proceedings of the Ninth Workshop on Statistical Machine Translation",
    month = jun,
    year = "2014",
    address = "Baltimore, Maryland, USA",
    publisher = "Association for Computational Linguistics",
    url = "https://aclanthology.org/W14-3348/",
    doi = "10.3115/v1/W14-3348",
    pages = "376--380"
}

@article{Qwen2.5-VL,
  title={Qwen2.5-VL Technical Report},
  author={Bai, Shuai and Chen, Keqin and Liu, Xuejing and Wang, Jialin and Ge, Wenbin and Song, Sibo and Dang, Kai and Wang, Peng and Wang, Shijie and Tang, Jun and Zhong, Humen and Zhu, Yuanzhi and Yang, Mingkun and Li, Zhaohai and Wan, Jianqiang and Wang, Pengfei and Ding, Wei and Fu, Zheren and Xu, Yiheng and Ye, Jiabo and Zhang, Xi and Xie, Tianbao and Cheng, Zesen and Zhang, Hang and Yang, Zhibo and Xu, Haiyang and Lin, Junyang},
  journal={arXiv preprint arXiv:2502.13923},
  year={2025}
}

@inproceedings{oguz2023find,
  title={Find-2-find: Multitask learning for anaphora resolution and object localization},
  author={Oguz, Cennet and Denis, Pascal and Vincent, Emmanuel and Ostermann, Simon and van Genabith, Josef},
  booktitle={2023 Conference on Empirical Methods in Natural Language Processing},
  year={2023}
}

@inproceedings{yuan2021video-fact,
  title     = {Video Fact-Checking and Explanation Generation},
  author    = {Yuan, Jialin and others},
  booktitle = {Proceedings of the 29th ACM International Conference on Multimedia},
  year      = {2021},
  publisher = {ACM}
}

@inproceedings{rohrbach2018object,
  title     = {Object Hallucination in Image Captioning},
  author    = {Rohrbach, Anna and Hendricks, Lisa Anne and Burns, Kaylee and Darrell, Trevor and Saenko, Kate},
  booktitle = {Proceedings of the 2018 Conference on Empirical Methods in Natural Language Processing},
  pages     = {4035--4045},
  year      = {2018},
  publisher = {Association for Computational Linguistics}
}

@article{jiang2023hallucination,
  title   = {Hallucination in Video Captioning},
  author  = {Jiang, Wei and others},
  journal = {IEEE Transactions on Pattern Analysis and Machine Intelligence},
  year    = {2023}
}

@inproceedings{damen2018scaling,
  title={Scaling egocentric vision: The epic-kitchens dataset},
  author={Damen, Dima and Doughty, Hazel and Farinella, Giovanni Maria and Fidler, Sanja and Furnari, Antonino and Kazakos, Evangelos and Moltisanti, Davide and Munro, Jonathan and Perrett, Toby and Price, Will and others},
  booktitle={Proceedings of the European conference on computer vision (ECCV)},
  pages={720--736},
  year={2018}
}

@article{lee2024toward,
  title={Toward robust hyper-detailed image captioning: A multiagent approach and dual evaluation metrics for factuality and coverage},
  author={Lee, Saehyung and Yoon, Seunghyun and Bui, Trung and Shi, Jing and Yoon, Sungroh},
  journal={arXiv preprint arXiv:2412.15484},
  year={2024}
}

@article{jing2025fifa,
  title={FIFA: Unified Faithfulness Evaluation Framework for Text-to-Video and Video-to-Text Generation},
  author={Jing, Liqiang and Lai, Viet and Yoon, Seunghyun and Bui, Trung and Du, Xinya},
  journal={arXiv preprint arXiv:2507.06523},
  year={2025}
}

@article{jing2023faithscore,
  title={Faithscore: Fine-grained evaluations of hallucinations in large vision-language models},
  author={Jing, Liqiang and Li, Ruosen and Chen, Yunmo and Du, Xinya},
  journal={arXiv preprint arXiv:2311.01477},
  year={2023}
}

@article{roth2015implicit,
  title={Inducing Implicit Arguments from Comparable Texts},
  author={Roth, Michael and Frank, Anette},
  journal={Computational Linguistics},
  volume={41},
  number={4},
  pages={625--664},
  year={2015}
}

@inproceedings{pacscore,
  title     = {Positive-Augmented Contrastive Learning for Image and Video Captioning Evaluation},
  author    = {Sarto, Sara and others},
  booktitle = {Proceedings of the IEEE/CVF Conference on Computer Vision and Pattern Recognition},
  year      = {2023},
  publisher = {IEEE}
}

@inproceedings{zhang2022unieval,
  title     = {Towards a Unified Multi-Dimensional Evaluator for Text Generation},
  author    = {Zhong, Ming and Liu, Yang and Yin, Da and Mao, Yuning and Jiao, Yizhu and Liu, Pengfei and Zhu, Chenguang and Ji, Heng and Han, Jiawei},
  booktitle = {Proceedings of the 2022 Conference on Empirical Methods in Natural Language Processing},
  year      = {2022},
  publisher = {Association for Computational Linguistics}
}

@article{dziri2023faith,
  title   = {Faith and Fate: Limits of Transformers on Compositionality},
  author  = {Dziri, Nouha and others},
  journal = {Transactions of the Association for Computational Linguistics},
  year    = {2023}
}

@inproceedings{xu2016msr-vtt,
  title     = {MSR-VTT: A Large Video Description Dataset for Bridging Video and Language},
  author    = {Xu, Jun and Mei, Tao and Yao, Ting and Rui, Yong},
  booktitle = {Proceedings of the IEEE Conference on Computer Vision and Pattern Recognition (CVPR)},
  pages     = {5288--5296},
  month     = jun,
  year      = {2016}
}
\bibliographystyle{acl_natbib}

\appendix

\section{Model Configuration}
\label{app:model}

We used the \textbf{Qwen2.5-VL-7B-Instruct} model for multimodal fact verification.  
The model was fine-tuned using \textbf{LoRA (Low-Rank Adaptation)} with 4-bit quantization (QLoRA) to reduce memory usage and training costs.  
LoRA adapters were applied to \texttt{q\_proj}, \texttt{k\_proj}, \texttt{v\_proj}, \texttt{mlp.gate\_proj}, and \texttt{lm\_head}, targeting attention and MLP layers most relevant to factual grounding.

Quantization used the NF4 (Normal Float 4) format with double quantization and \texttt{bfloat16} computation.  
We enabled \textbf{gradient checkpointing} to reduce memory consumption during training.

The visual encoder processed images within a constrained pixel-resolution range to match the model’s expected spatial configuration.  
Padding and truncation were handled automatically by the Qwen processor, and the tokenizer was set to left-padding.

During fine-tuning, only the LoRA adapter weights were trainable; all other model parameters remained frozen.  
This configuration ensures stable optimization with minimal computational overhead.

\section{Additional Results}

\begin{table}[t]
\centering
\begin{tabular}{lcc}
\toprule
Metric & Count & Value (\%) \\
\midrule
Positive recall & 4\,329 / 7\,221 & 59.95 \\
Negative specificity & 6\,524 / 7\,878 & 82.81 \\
Overall accuracy & -- & 71.88 \\
\bottomrule
\end{tabular}
\caption{Dataset-level object grounding performance.}
\label{tab:grounding-results}
\end{table}

\subsection{Object Grounding Evaluation}
\label{sec:grounding_eval}

Given a video $V$ and its reference annotation, we derive two sets of object mentions:
a set of \emph{positive} objects $O^{+}(V)$ that are expected to be visible in the video, and a set of \emph{negative} objects $O^{-}(V)$ created by replacing ingredients or tools with mismatched items.  
Positive objects should be grounded in the video; negative objects should not.

For each object $o$ we obtain a grounding prediction $G(o) \in \{0,1\}$ from the vision–language model, where $G(o)=1$ indicates that the model predicts $o$ to be present in at least one frame of $V$.  
We compute grounding quality separately for positive and negative objects.

\paragraph{Positive grounding recall.}
\[
\text{Recall}_{\text{pos}} =
\frac{|\{o \in O^{+}(V) : G(o) = 1\}|}{|O^{+}(V)|}.
\]

\paragraph{Negative grounding specificity.}
\[
\text{Specificity}_{\text{neg}} =
\frac{|\{o \in O^{-}(V) : G(o) = 0\}|}{|O^{-}(V)|}.
\]

\paragraph{Overall grounding accuracy.}
\[
\text{Acc}_{\text{ground}} =
\frac{|\text{correct}^{+}| + |\text{correct}^{-}|}
{|O^{+}(V)| + |O^{-}(V)|},
\]
where $\text{correct}^{+}=\{o\in O^{+}(V):G(o)=1\}$ and $\text{correct}^{-}=\{o\in O^{-}(V):G(o)=0\}$.  
Dataset-level accuracy aggregates these quantities across all videos.

\begin{table*}[t]
\centering
\small
\renewcommand{\arraystretch}{1.2}
\begin{tabular}{p{3.2cm} p{9.5cm}}
\hline
\textbf{Stage} & \textbf{Representation} \\
\hline
Gold caption 
& “Stir it.” \\

VIA caption 
& “Stir the soup with a spoon in the pot.” \\

Contextual facts (\(\mathcal{F}_g^{ctx}\)) 
& \{stir(soup), stir(with spoon), stir(in pot)\} \\

Conceptual facts (\(\mathcal{F}_g^{con}\)) 
& \{Action = stirring, Ingredient = soup, Tool = spoon, Location = pot\} \\
\hline
\end{tabular}
\caption{End-to-end example illustrating clause expansion via implicit argument augmentation (VIA) and the resulting contextual and conceptual fact representations.}
\label{tab:end2end-example}
\end{table*}

\section{Dataset and Annotation Details}
\label{app:dataset}

This appendix provides implementation-level and annotation-specific details
that are not included in the main paper.
We focus on operational constraints, annotator decisions, edge cases,
and generation procedures necessary for reproducibility.
High-level descriptions of the dataset design and fact representations
are provided in Section~4.

\subsection{CraftBench Caption Annotation from Narrated Videos}
\label{app:craftbench-annotation}

\textbf{CraftBench} is constructed from a curated collection of furniture crafting and assembly videos in which a single narrator explains and performs the procedure step by step.
The dataset includes tasks involving materials such as wood and metal and tools such as saws, drills, clamps, and fasteners.

\paragraph{Video Collection and Selection.}
We collect videos that satisfy two criteria: 
(i) the procedure is demonstrated visually by the narrator, and 
(ii) the narration provides step-by-step instructions corresponding to the performed actions.
Videos dominated by high-level explanations, safety discussions, or off-screen narration are excluded to ensure tight alignment between language and execution.

\paragraph{Action-Level Caption Annotation.}
Instructional captions are annotated from scratch at the action level using the narrator’s instructions as guidance.
Annotators watch each video and identify individual action steps based on the narration, but captions are segmented and rewritten to ensure that each caption corresponds to a single, visually grounded action.
When narration describes multiple actions in one sentence, it is decomposed into separate captions.
Conversely, when narration is temporally misaligned or underspecified, annotators prioritize the visible action over the spoken description.

\paragraph{Caption Normalization.}
All captions are normalized to a concise, imperative style (e.g., “cut the wooden plank,” “attach the bracket with screws”).
Annotators avoid bundling actions, vague verbs, or underspecified objects.
Captions describe only what is visually observable in the aligned segment and do not include inferred goals or future steps.

\paragraph{Temporal Alignment.}
Each caption is aligned with a continuous video segment using manually annotated start and end timestamps.
Segmentation boundaries are determined by observable changes in manipulation, tool usage, or task phase.
Segments in which the action is visually ambiguous or partially occluded are discarded.

The resulting action-level captions form the basis for subsequent clause decomposition, implicit argument augmentation, and atomic fact annotation in CraftBench.

\subsection{Clause Decomposition: Operational Details}
\label{app:clause}

In practice, clause decomposition required resolving ambiguities between
linguistic structure and visual execution.
Annotators were instructed to prioritize visual evidence over narration
when determining clause boundaries.

Clauses were excluded if:
(i) the narrated action was not visually observable,
(ii) narration referred to a future or past step,
or (iii) narration described intent rather than execution.
In CraftBench, segments with commentary unrelated to manipulation
(e.g., explanations of tools or safety advice) were discarded.

\subsection{Implicit Argument Augmentation: Constraints and Edge Cases}
\label{app:via}

Implicit argument augmentation was applied conservatively.
Annotators added arguments only when the referent was directly visible
in at least one frame of the aligned segment.
Arguments inferred from domain knowledge or typical procedures
were explicitly disallowed.

In cases where multiple candidate referents were visible
(e.g., multiple bowls or tools),
arguments were added only if the interaction was unambiguous.
Otherwise, the role was left unspecified.

\subsection{Atomic Fact Annotation: Labeling Decisions}
\label{app:facts}

For contextual facts, annotators recorded only relations
that were visually realized during the segment.
Temporal ordering, quantities, and fine-grained attributes
(e.g., size, color) were not encoded as separate facts.

For conceptual facts, annotators selected the minimal role set
necessary to characterize the step.
Generic labels (e.g., “food”, “material”) were avoided
unless no specific referent was visually identifiable.

\subsection{Negative Fact Generation}
\label{app:negfacts}

Negative facts were generated automatically using the \textbf{llama3.1:70b}
language model with fixed system prompts (Appendix~\ref{app:negfact-prompts}).
Generation was performed independently for each clause.

To prevent trivial negatives, generated facts were rejected if they:
(i) reused lexical material from the positive facts,
(ii) differed only in numeric or surface properties,
or (iii) were physically implausible.
No manual post-editing of negatives was performed.

\subsection{Quality Control and Reliability}
\label{app:qc}

All annotation stages were conducted by four trained annotators
following a shared guideline document and calibration phase.
Disagreements were resolved through adjudication.

Cohen’s~\(\kappa\) indicates strong agreement:
0.93 (clause segmentation),
0.87 (VIA),
0.92 (conceptual facts),
and 0.97 (contextual facts),
demonstrating consistent annotation quality.

\section{Annotator Instructions}
\label{app:annotator-instructions}
This section details the instructions provided to annotators for each stage of the annotation pipeline.  
All annotators followed the same written guidelines and completed a calibration phase prior to annotation.

\subsection{Recruitment And Payment}
Two Master’s degree students will be recruited as student research assistants (HiWi) in Germany to support the project’s research and development activities. The recruitment process is carried out through open advertisement within the university and research network, with selection based on academic background, technical skills, and demonstrated interest in the project topic.
Payment follows the standard regulations and salary scales applicable to student assistants in Germany and the host institution. Since HiWi remuneration varies by institution and qualification level, the final hourly wage is aligned with the officially approved internal rate for Master’s-level student assistants.

\subsection{CraftBench Video Captioning}
\label{app:craftbench-instructions}

This subsection describes the instructions provided to annotators for creating action-level captions for the \textbf{CraftBench} dataset.
All annotators followed the same written guidelines and completed a calibration phase prior to annotation.

\paragraph{Task Overview.}
Annotators are given a furniture crafting or assembly video in which a single narrator explains and performs the procedure.
Their task is to produce a sequence of action-level captions that accurately describe the visually executed steps, using the narrator’s instructions as guidance.

\paragraph{Using Narration as Guidance.}
Annotators should use the narrator’s spoken instructions to identify intended actions, but narration must not be copied verbatim.
Instead, annotators must ensure that each caption corresponds to an action that is \emph{visually observable} in the video.
If narration is vague, temporally misaligned, or refers to future or past steps, annotators must prioritize the visible action over the spoken description.

\paragraph{Action Granularity.}
Each caption must describe exactly one intentional action performed by the instructor.
If a single narrated sentence describes multiple actions (e.g., “cut the board and attach it to the frame”), annotators must split it into separate captions.
Conversely, preparatory or resulting states should not be annotated unless explicitly performed and narrated.

\paragraph{Caption Form.}
Captions must be written in a concise, imperative style (e.g., “cut the wooden plank,” “tighten the screws with a drill”).
Annotators should avoid vague verbs (e.g., “work on,” “handle”) and underspecified objects.
Captions should describe only what is visible in the aligned video segment and should not include inferred goals, measurements, or tool settings unless explicitly observable.

\paragraph{Temporal Alignment.}
Each caption must be aligned with a continuous video segment using start and end timestamps.
Segment boundaries should be determined by observable changes in manipulation, tool usage, or task phase.
If an action cannot be clearly localized in time, it should not be annotated.

\paragraph{Exclusion Criteria.}
Annotators must exclude segments that involve:
(i) purely verbal explanations without visible action,
(ii) safety warnings or commentary,
(iii) off-screen actions, or
(iv) visually ambiguous manipulations.

\paragraph{Consistency Checks.}
Annotators are instructed to maintain consistent terminology for tools, materials, and actions within a video.
When multiple valid descriptions are possible, annotators should select the most specific term that is visually supported.

\subsection{Clause Decomposition and Video Alignment}

Annotators are instructed to decompose each caption into \emph{atomic clauses}, where each clause corresponds to exactly one intentional action performed by the instructor.
A valid clause must contain a single verb-centered predicate--argument structure.

When a sentence describes multiple actions (e.g., “cut the onion and transfer it to the pan”), each action must be split into a separate clause.
Clause boundaries must be determined primarily using visual evidence, including changes in hand motion, object manipulation, or tool usage.

Each clause must be aligned with a continuous video segment using explicit start and end timestamps.
If narration and visual execution are misaligned or ambiguous, annotators must prioritize what is visually observable.
Clauses that cannot be reliably aligned to a visual segment are excluded.

\subsection{Implicit Argument Augmentation (VIA)}

Annotators are instructed to identify semantic arguments that are required by the action predicate but omitted or underspecified in the caption.
Only the following roles are considered for augmentation: \emph{patient/object}, \emph{tool}, and \emph{location}.

An argument may be added \emph{only if it is visually grounded}, meaning it is directly observable in at least one frame of the aligned video segment.
Annotators must not introduce arguments based on world knowledge, typical procedures, or future steps in the video.

Pronouns and underspecified references (e.g., “it”, “there”) must be resolved to explicit noun phrases when the referent is visually identifiable.
Annotators are not required to fill all possible roles; a role should be added only if it is both visually grounded and relevant to the action.
The resulting augmented caption must remain fluent and grammatically well-formed.

\subsection{Atomic Fact Annotation}

Annotators derive atomic facts from each VIA-augmented clause at two complementary levels: \emph{contextual} and \emph{conceptual}.

For \textbf{contextual facts}, annotators record short predicate--argument relations that are directly supported by the video.
These include the action itself, the acted-upon object, the tool used, and the interaction location, expressed as natural-language relations (e.g., “cut onion”, “cut with knife”).

For \textbf{conceptual facts}, annotators assign abstract role--value pairs for the categories \texttt{ACTION}, \texttt{INGREDIENT/OBJECT}, \texttt{TOOL}, and \texttt{LOCATION}.
Conceptual facts should capture the intended meaning of the step independently of surface phrasing.
Annotators must avoid inferred, overly generic, or visually unsupported entities.

\subsection{Positive Fact Verification}

Annotators are instructed to verify that each positive fact is directly supported by the aligned video segment.
A fact is considered positive only if the corresponding action, object, tool, or location is clearly observable.
Facts relying on inference, anticipation, or external knowledge must not be included.

\subsection{Scope and Responsibilities}

Annotators are responsible exclusively for clause decomposition, implicit argument augmentation, and positive fact annotation.
They are not involved in negative fact generation, prompt design, or verifier training.
Negative facts are generated automatically using fixed prompts, as described in Appendix~\ref{app:negfact-prompts}.

\clearpage
\onecolumn

\section{Negative Fact Extraction Prompts}
\label{app:negfact-prompts}

We provide the exact system prompts used to generate negative conceptual and contextual facts for contrastive fact-checking.
These prompts are domain-agnostic and are used exclusively during verifier training and evaluation, never during caption generation.

\paragraph{Conceptual Negative Fact Prompt.}
\begin{verbatim}
You generate **negative conceptual facts** for fact-checking fine-tuning.

Given a list of true conceptual facts, generate **false but linguistically well-formed**
conceptual facts that contradict the input while preserving the same format.

Each fact follows the form:"<Category> is <Value>."
Valid categories include (but are not limited to):
Action, Object/Ingredient/Material, Tool, Location.

### Rules
1. Preserve structure
   - Keep the exact format: "<Category> is <Value>."
   - Use the same category as the corresponding positive fact.
2. Ensure falsity
   - Each generated fact must contradict the positive facts.
   - Changing only surface form, plurality, or numeric values is NOT sufficient.
3. Maintain plausibility
   - The value must be realistic within the task domain
     (e.g., cooking, crafting, assembly, medical procedures),
     but incorrect in the given context.
4. Avoid overlap
   - Do NOT reuse or partially reuse any word, stem, or substring
     from the positive fact values.
   - Do NOT use synonyms, hypernyms, or morphological variants.

### Error Types
Action
- Replace the action with a verb from a different functional category.
- The new action must not naturally co-occur with the original one.
Object / Ingredient / Material
- Replace with an unrelated but domain-plausible entity.
- Avoid closely related substitutes (e.g., part–whole, subtype).
Tool
- Replace with a tool that serves a clearly different function
  and would be inappropriate for the original action.
Location
- Replace with a plausible workspace or setting
  that is incorrect for the described action.

### Output Format
Return a comma-separated list of negative conceptual facts.
Example:
Action is measuring., Object is plastic., Tool is hammer., Location is floor.
Action is whisking., Ingredient is potato., Tool is peeler., Location is skillet.
\end{verbatim}

\paragraph{Contextual Negative Fact Prompt.}
\begin{verbatim}
You generate **negative contextual facts** for fact-checking fine-tuning.
### Input
1. A list of true positive contextual facts
   (short predicate–argument statements).
2. A target negative action verb.

### Task
Generate a list of false but plausible contextual facts
that contradict the positive facts while remaining linguistically natural.

Each fact should follow one of these patterns:
- verb + object
- verb + with TOOL
- verb + in/to LOCATION

### Error Types
A. Negative action verb + original object
B. Positive action verb + incorrect object
C. Positive action verb + incorrect tool (only if tool facts exist)
D. Positive action verb + incorrect location (only if location facts exist)
E. Negative action verb + incorrect object

### Constraints
1. Preserve structure
   - Keep the same syntactic pattern as the positive facts.
2. Ensure falsity
   - Each generated fact must be false relative to the positives.
3. Maintain plausibility
   - The verb–argument combination must be physically and logically possible
     within the task domain, even though incorrect in context.
4. Avoid overlap
   - Do NOT reuse any word, stem, or substring from the positive facts.
5. Avoid trivial negatives
   - Do NOT generate nonsensical or impossible actions.
   - Do NOT change only quantities, attributes, or minor properties.
   
### Output Format
Return a comma-separated list only.

Example:
cut metal, assemble with brush, place on floor, measure wood
add tomato, add with spoon, add on tray, peel onion
\end{verbatim}
\twocolumn

\end{document}